\documentclass{article}

\usepackage{arxiv}

\usepackage[utf8]{inputenc} 
\usepackage[T1]{fontenc}    
\usepackage{hyperref}       
\usepackage{url}            
\usepackage{booktabs}       
\usepackage{amsfonts}       
\usepackage{nicefrac}       
\usepackage{amsmath}
\usepackage{microtype}      
\usepackage{cleveref}       
\usepackage{lipsum}         
\usepackage{graphicx}
\usepackage[numbers]{natbib}
\usepackage{doi}
\usepackage{multirow}
\usepackage{xcolor}         

\title{FB-HyDON: Parameter-Efficient Physics-Informed Operator Learning of Complex PDEs via Hypernetwork and Finite Basis Domain Decomposition}

\newif\ifuniqueAffiliation
\uniqueAffiliationtrue

\ifuniqueAffiliation 
\author{Milad Ramezankhani\thanks{Corresponding author. Email: milad.ramezankhani@quantiphi.com} \\
	Applied Research, Quantiphi\\
	Marlborough, MA 01752, USA \\
	\And
        Rishi Yash Parekh \\
	Applied Research, Quantiphi\\
	Marlborough, MA 01752, USA \\
	\AND
        Anirudh Deodhar \\
	Applied Research, Quantiphi\\
	Marlborough, MA 01752, USA \\
	\And
        Dagnachew Birru \\
	Applied Research, Quantiphi\\
	Marlborough, MA 01752, USA \\
}
\else
\usepackage{authblk}

\newbox{\orcid}\sbox{\orcid}{\includegraphics[scale=0.06]{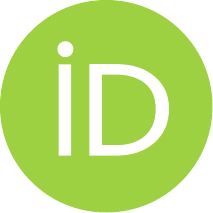}} 
\author[1]{%
	\href{https://orcid.org/0000-0000-0000-0000}{\usebox{\orcid}\hspace{1mm}David S.~Hippocampus\thanks{\texttt{hippo@cs.cranberry-lemon.edu}}}%
}
\author[1,2]{%
	\href{https://orcid.org/0000-0000-0000-0000}{\usebox{\orcid}\hspace{1mm}Elias D.~Striatum\thanks{\texttt{stariate@ee.mount-sheikh.edu}}}%
}
\affil[1]{Department of Computer Science, Cranberry-Lemon University, Pittsburgh, PA 15213}
\affil[2]{Department of Electrical Engineering, Mount-Sheikh University, Santa Narimana, Levand}
\fi

\renewcommand{\shorttitle}{FB-HyDON for Operator Learning of Complex PDEs}

\hypersetup{
pdftitle={FB-HyDON for operator learning of complex PDEs},
pdfsubject={},
pdfauthor={Milad Ramezankhani, Rishi Yash Parekh, Anirudh Deodhar, Dagnachew Birru},
pdfkeywords={Operator learning, physics-informed neural networks, hypernetwork, domain decomposition},
}

\begin{document}
\maketitle

\begin{abstract}
    Deep operator networks (DeepONet) and neural operators have gained significant attention for their ability to map infinite-dimensional function spaces and perform zero-shot super-resolution. However, these models often require large datasets for effective training. While physics-informed operators offer a data-agnostic learning approach, they introduce additional training complexities and convergence issues, especially in highly nonlinear systems. To overcome these challenges, we introduce  Finite Basis Physics-Informed HyperDeepONet (FB-HyDON), an advanced operator architecture featuring intrinsic domain decomposition. By leveraging hypernetworks and finite basis functions, FB-HyDON effectively mitigates the training limitations associated with existing physics-informed operator learning methods. We validated our approach on the high-frequency harmonic oscillator, Burgers' equation at different viscosity levels, and Allen-Cahn equation demonstrating substantial improvements over other operator learning models. 
\end{abstract}


\section{Introduction}
Partial differential equations (PDEs) are integral in modeling and describing the dynamics of many complex systems in science and engineering. Numerical solvers such as finite element methods (FEMs) and finite difference methods (FDMs) often obtain the solution of PDEs by discretizing the domain and solving a finite-dimensional problem. However, obtaining high-resolution solutions to PDEs using numerical simulations for complex large-scale problems can be computationally expensive and prohibitive. There has been a growing interest in more efficient data-driven alternatives that can directly learn the underlying solutions from the available data without requiring explicit knowledge about the governing PDEs \cite{belbute2020combining, khoo2021solving}. More recently, operator learning has emerged as a promising paradigm, aiming to learn an unknown mathematical operator governing a system of PDEs \cite{boulle2023mathematical}. They capture mappings between \textit{infinite-dimensional} function spaces and have demonstrated potential in capturing complex solution behaviors \cite{lu2021learning, li2020fourier}. Furthermore, due to their inherent differentiability, they can be seamlessly applied to inverse problems, such as design optimization tasks \cite{azizzadenesheli2024neural}.  Various architectures have been developed, including the Deep Neural Operator (DeepONet) \cite{lu2021learning}, Fourier Neural Operator (FNO) \cite{li2020fourier}, Graph Neural Operator \cite{li2020neural}, General Neural Operator Transformer (GNOT) \cite{hao2023gnot} and Operator Transformer (OFormer) \cite{li2022transformer}. These models differ in their discretization methods and the approximation techniques they use to enhance efficiency and scalability. 

Training operators, however, relies on large datasets, which may not be readily available for many practical applications and can result in suboptimal generalization performance. One way to reduce (or eliminate) the operators' data dependency is to augment the training process with physical laws and learn the operator in a physics-informed fashion \cite{wang2021learning, li2024physics, ramezankhani2024advanced}. Despite their advantages, such physics-informed models often face challenges related to complex optimization landscapes and convergence difficulties, hindering their effectiveness \cite{krishnapriyan2021characterizing}. 
In this paper, we introduce a novel operator-learning architecture designed to overcome the aforementioned limitations. We present the Finite Basis Physics-Informed HyperDeepONet (FB-HyDON), which efficiently and accurately learns the solution operator for complex, highly nonlinear, and high-frequency problems. The primary contributions of this study are as follows: 1) We propose FB-HyDON, an advanced operator architecture with built-in domain decomposition functionality that effectively addresses the training limitations of existing physics-informed operators; 2) We introduce a hypernetwork-based variant of finite basis domain decomposition method \cite{moseley2023finite} which facilitates learning complex PDE systems in a more parameter-efficient manner and maintains a constant number of trainable parameters for any domain decomposition, from coarse to fine; 3) We conduct a comparative analysis of FB-HyDON against various operator models, demonstrating that our model consistently achieves superior results while incorporating only a fraction of the model complexity needed in other operator methods.

\section{Background}
\textbf{Operator learning}. Considering a parametric PDE taking the form \(\mathcal{N}(u,a)=0\) where \(u \in \mathcal{U}\) is the unknown solution, \(a \in \mathcal{A} \subseteq \mathcal{V} \) is a PDE parameter and \(\mathcal{N}:\mathcal{U}\times \mathcal{A} \rightarrow \mathcal{F}\) is a linear or nonlinear partial differential operator with \(\mathcal{U}, \mathcal{V}, \mathcal{F}\) representing a triplet of Banach spaces. Given suitable initial and boundary conditions, for any \(a \in \mathcal{A} \), we assume that there is a unique solution \(u = u(a)\) to the problem. This formulation results in the solution operator \(\mathcal{G}: \mathcal{A} \rightarrow \mathcal{U}\). The solution operator \(\mathcal{G}\) can be approximated by an operator \(\mathcal{G_\theta}\) with parameters \(\theta\) \cite{li2024physics}. Specifically in DeepONet's architecture, \(\mathcal{G_\theta}\) comprises a branch net and a trunk net.  The branch net encodes the input function information at discrete sensor points and the trunk net embeds the spatiotemporal coordinates of the PDE system. The networks' outputs are then merged through an element-wise multiplication followed by a summation to generate the PDE solution \(\mathcal{G_\theta}(a)(y)\)
\cite{lu2021learning}.

\textbf{HyperDeepONet}. Since DeepONet approximates the target functions with a finite-dimensional linear subspace, it struggles to accurately learn complex functions with nonlinear and sharp features \cite{hadorn2022shift}.  Various works have been proposed to address this limitation \cite{hadorn2022shift, seidman2022nomad, ramezankhani2024advanced}. Recently, HyperDeepONet (HDON) \cite{lee2023hyperdeeponet} has been introduced as a more expressive variant of DeepONet capable of learning highly nonlinear operators with fewer network parameters. HDON replaces the branch net with a hypernetwork which infers the parameters of the trunk (target) net. Instead of propagating the input function information only through the last layer of the branch and trunk nets, HDON infuses this information into every parameter of the trunk net, enabling a more comprehensive integration of the input function across the entire network.

\textbf{Physics-informed operator learning and domain decomposition}. Physics-informed operator learning allows learning the operator using only the form of the PDE and its initial and boundary conditions, without the need for large datasets. By embedding the governing physical laws as constraints within the training process, operating learning methods can be trained in a fully data-agnostic \cite{lu2021learning}, or hybrid manner \cite{li2024physics}. However, similar to physics-informed neural networks (PINNs) \cite{raissi2019physics}, physics-informed operator learning is not without limitations. One major challenge is the convergence issues when dealing with complex systems, such as those with nonlinear time-varying characteristics and sharp transitions, which can severely impact the performance \cite{krishnapriyan2021characterizing, wang2023expert, wang2024respecting}. Domain decomposition, an effective way to tackle such challenges, involves breaking down the PDE domain into smaller, more manageable subproblems, and thus enhancing convergence. Two notable approaches in this context are extended PINN (XPINN) \cite{jagtap2020extended} and finite basis PINN (FBPINN). FBPINN \cite{moseley2023finite} is particularly interesting since, unlike XPINN, it does not introduce new loss terms, thereby maintaining a simpler optimization landscape.

\section{Methodology}
\subsection{FB-HyPINN} \label{FBHyPINN}

In FBPINN's framework, the input domain $\Omega$ is partitioned into $J$ overlapping subdomains such that $\Omega = \bigcup_{j=1}^{J} \Omega_j$ and the intersection of any two adjacent subdomains is non-empty, i.e., $\Omega_j \cap \Omega_k \neq \emptyset$ for $k$ being an adjacent subdomain to $j$. A family of sub-networks, denoted as $V$, is defined as: 
\begin{align}
      V = \{ \hat{u}_{j}(x,\theta_{j}) \mid x \in \Omega, \theta_{j} \in \Theta_{j} \}_{j=1}^{J}  
\end{align}
  
\begin{figure}[ht] 
    \centering
    \includegraphics[width=\linewidth]{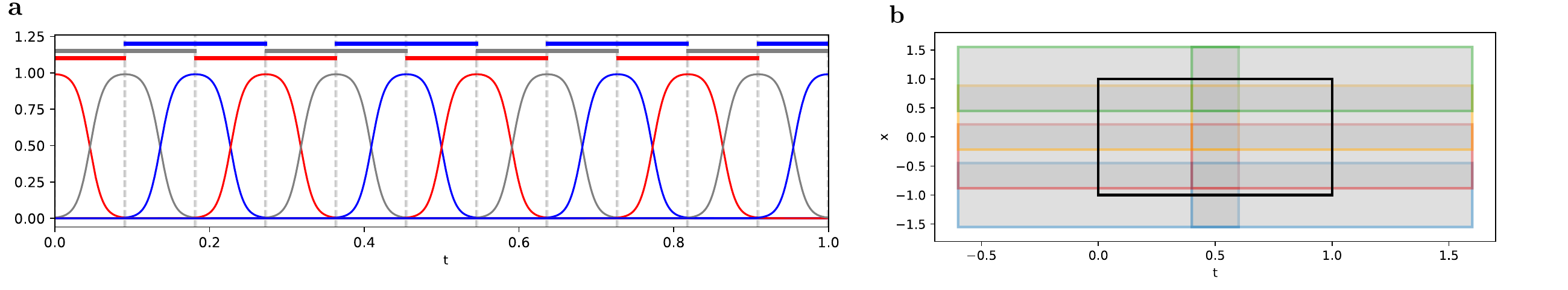} 
    \caption{a) Visualization of a set of window functions for harmonic oscillator case where  time domain is divided into 12 subdomains.  Different colors represent different subdomains and all subdomains are associated with window function based on their bounds. X-axis represents the time domain and y-axis shows window function values. Horizontal bars on the top represent the length of subdomains. b) Visualization of domain decomposition for time-space domain of 1D Allen-Cahn example. Black box represents the problem domain and overlapping subdomains are shown with gray rectangles.}
    \label{fig:window_function}
\end{figure}

Each subnetwork \(\hat{u}_{j}(x,\theta_{j})\) corresponds to a specific subdomain $\Omega_{j}$. These subnetworks operate independently to learn the solution relevant to their assigned subdomain. According to the FBPINN scheme \cite{moseley2023finite}, the inputs to the subnetworks are normalized separately over each subdomain, which mitigates the issue of spectral bias. For every collocation point $x_i$, the outputs of the subnetworks are generated and multiplied with a window value $\omega_{j}(x_i)$ determined by their respective window function and summed across all subdomains. An example and description of the window functions and subdomains are provided in Figure \ref{fig:window_function}. The primary role of window functions is to bound each subnetwork to its subdomain by introducing a higher weight near the center of the corresponding subdomain and zero outside of it.  

In our proposed method, FB-HyPINN, we utilize a hypernetwork $\hat{h}( d,\theta_{H})$ \cite{ha2016hypernetworks} that computes the parameters $\{\theta_{j} \}_{j=1}^{J}$ of the subnetworks $\{\hat{u}_{j}\}_{j=1}^{J}$. The input $d$ to the hypernetwork is a vector that uniquely defines a subdomain. In our approach we pick the subdomain's midpoint $m_{j}$ and the distance $s_{j}$ between the midpoint and subdomain's bounds as the input to the hypernetwork. Hence, the output of each subnetwork \(\hat{u}_{j}(x_i^{norm}, \theta_{j})\) is calculated via obtaining the parameters from the hypernetwork \(\theta_j = \hat{h}((m_{j},s_{j}), \theta_{H})\) and feeding in the normalized input variables \(x_i^{norm} = Norm(x_{i}, \Omega_{j}).\) Subsequently, the global solution $\hat{u}(x_{i}, \theta_{H})$  for a collocation point $x_{i}$ is obtained as
\begin{equation}
    \hat{u}(x_{i}, \theta_{H}) = \sum_{j=1}^{J} \omega_{j}(x_i)\hat{u}_{j}(x_i^{norm}, \theta_{j}).
\end{equation}

In a given PDE of the form
    \begin{align}
        &\mathcal{N}[u](x) = 0 ,  x \in \Omega\\
        &u(x) = g(x), x \in \partial\Omega,
    \end{align}

$\mathcal{N}[.]$ denotes a differential operator, $u(x): \overline{\Omega} \rightarrow \mathbb{R^{d}}$ is the unknown solution, \(\ x = \{x_i\}_{i=1}^{N_I} \) denotes a collection of collocation points sampled within the domain interior and \(\{x_{j}\}_{j=1}^{N_{B}}\) represents a set of points sampled along each boundary condition. Additionally, \(\lambda_I\) and \(\lambda_{B}\) are appropriately selected scalar weights. Just like a regular PINN framework, here, the solution \(u(x)\) is approximated by $\hat{u}(x_{i}; \theta_{H})$ via the defined physics and boundary losses and optimizing the hypernetwork parameters $\theta_{H}$. The loss is represented as:
    \begin{align}
        &\mathcal{L}(\theta_{H}) = \frac{\lambda_{I}}{N_{1}}\sum_{i=0}^{N_{1}}(\mathcal{N}[\hat{u}(x_{i}, \theta_{H})])^2 + \frac{\lambda_{B}}{N_{B}}\sum_{i=0}^{N_{B}}(\hat{u}(x_{i}, \theta_{H}) - g(x_{i}))^2.
    \end{align}

\subsection{FB-HyDON}
\begin{figure}[ht] 
    \centering
    \includegraphics[width=\linewidth]{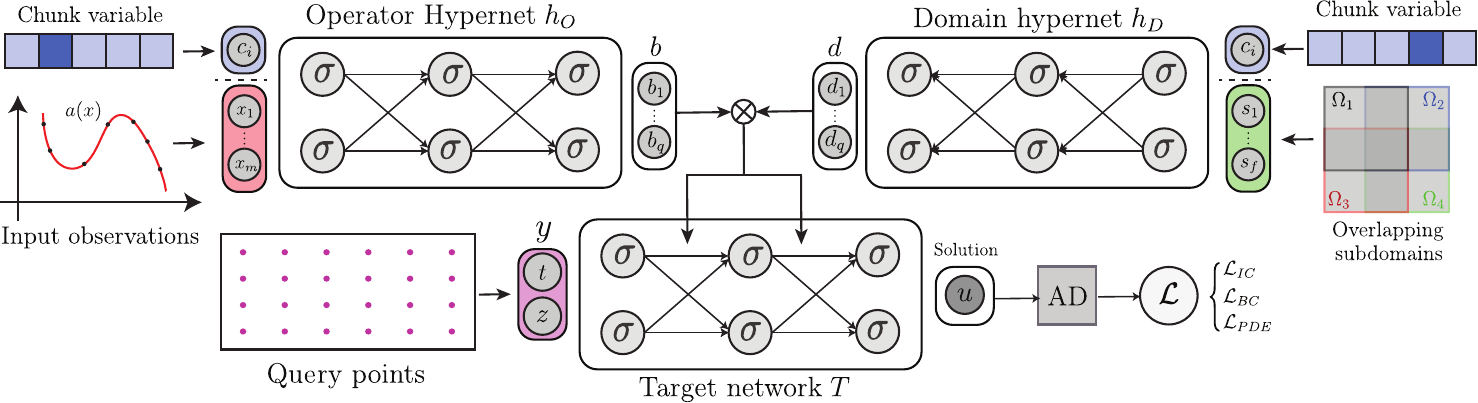}  
    \caption{Schematic of proposed FB-HyDON model. The inputs to the hypernetworks consist of a chunk identifier and a task-specific set of variables (sensory observations for Operator hypernet and subdomain information for Domain hypernet.) The outputs of the hypernets are merged together via the Hadamard product to produce the weights of the Target net. Physics-informed losses are obtained at query points and used to train the hypernets' parameters.}
    \label{fig:model}
\end{figure}

\textbf{Dual-hypernetwork module}. As depicted in Figure \ref{fig:model}, the proposed FB-HyDON model has two hypernetworks with equal output dimensions followed by a target network which generates the output solution for any given query point \(y\) on the spatiotemporal domain. Operator hypernet \(h_O\) is responsible for mapping the input function observations \([a(x_1), a(x_2), \ldots, a(x_m)] \) at sensor points to a feature representation \([b_1, b_2, ..., b_q]^T\). Domain hypernet \(h_D\) on the other hand takes in the subdomain coordinates \([s_1, s_2, ..., s_f]^T\) as the input and generates a feature embedding \(d\). The hypernets' outputs are then merged via the Hadamard product, generating the weights of the target network. While one hypernet might be sufficient to take in and process both input functions and subdomain information, we observed that separate hypernets considerably improve the model's overall performance.

\textbf{Finite basis domain decomposition with hypernetwork}. FBPINN partitions the problem domain into smaller, overlapping subdomains, each associated with a dedicated subnetwork that learns a basis function with compact support \cite{moseley2023finite}. The overall solution is constructed by combining the basis functions (analogous to FEMs), where the prediction for each query point is achieved by the weighted sum of outputs from subnetworks whose corresponding subdomains encompass that point. However, this approach presents a significant drawback: the number of subnetworks (i.e., trainable parameters) increases with the number of subdomains, leading to potential inefficiencies in training. Rather than employing separate subnetworks for each subdomain, we propose FB-HyPINN, a novel strategy utilizing a hypernetwork architecture. The hypernetwork (\(h_D\) in Figure \ref{fig:model}) generates the parameters for each subdomain's network, reducing the overall parameter count while maintaining model performance. Each subdomain is encoded into a unique vector 
\(s\) that captures its spatial information. This vector serves as input to the hypernetwork, which generates the embedding \(d\) for the corresponding subnetwork. The embedding is then used to generate the target network's parameters predicting the operator output.

\textbf{Chunked hypernetwork}. As the size of the target network increases, the output layer of the hypernetworks can get prohibitively large, potentially hindering both training efficiency and model performance. \textit{Chunked hypernetworks} \cite{von2019continual} offer an effective solution to this challenge. In this method,  the parameters of the target network are generated in smaller, manageable chunks over multiple steps via iteratively invoking the hypernetwork. To differentiate between chunks, an additional input \(\mathcal{C} = \{c_i\}_{i=1}^{N_c}\)  is introduced to the hypernetwork which allows generating chunk-specific outputs for a given fixed task \(t\). The full set of target network parameters can then be obtained by concatenating the outputs of the hypernetwork for each chunk: \(\theta_{t} = \left[ h(t, c_1), \dots, h(t, c_{N_c}) \right] \). In the proposed framework, both hypernetworks (\(h_O\) and \(h_D\)) leverage the chunking strategy (Figure \ref{fig:model}).

\section{Results}
\subsection{Physics-informed learning with finite basis domain decomposition}
\begin{figure}[h] 
    \centering
    \includegraphics[width=0.7\linewidth]{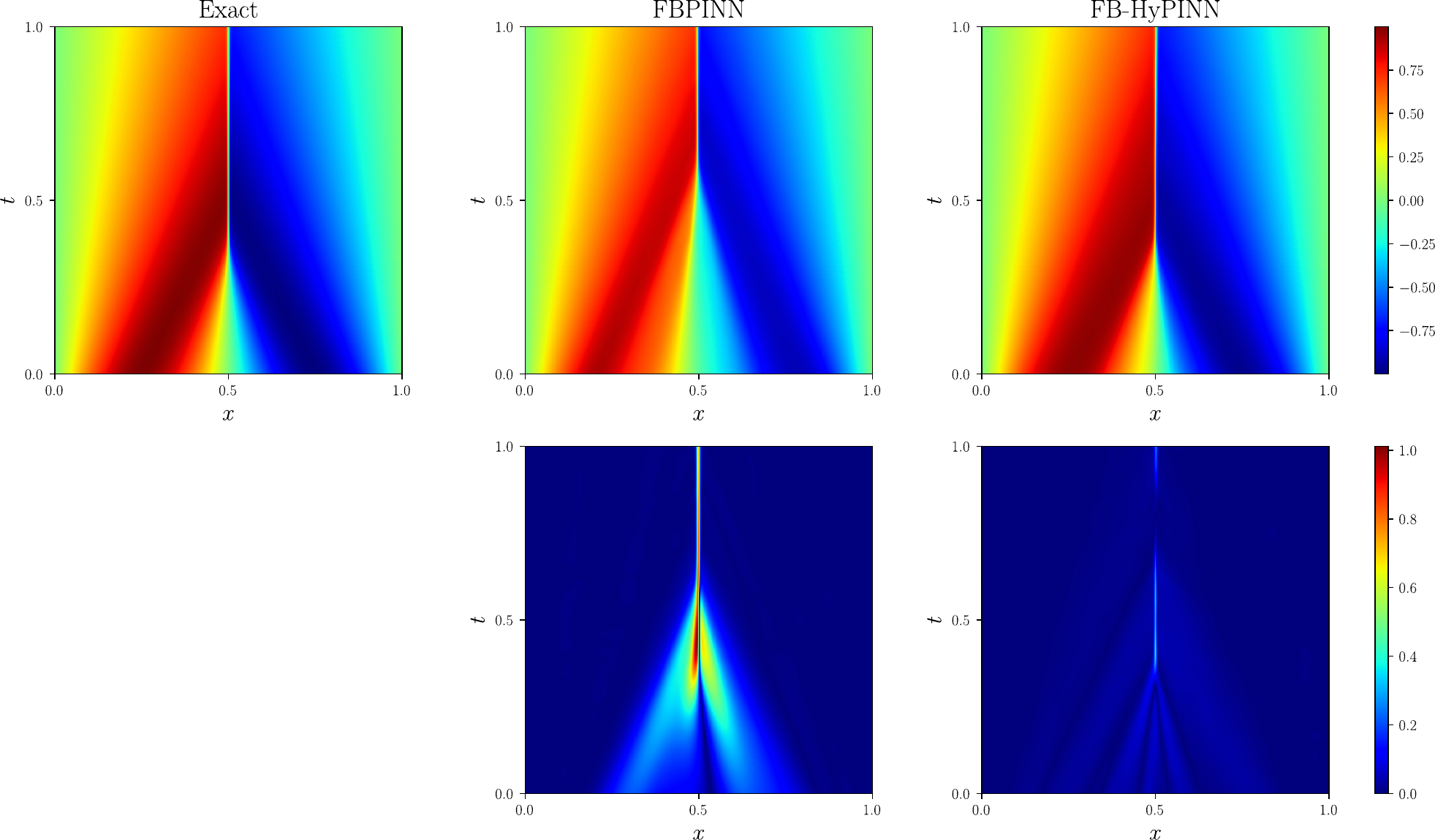} 
    \caption{Comparison of FBPINN and FB-HyPINN's predictions (top row) and absolute error (bottom row) for solving 1D Burger's equation with \(\nu = 0.001/\pi \) via physics-informed learning with domain decomposition. Both model has 25 subdomains.}
    \label{fig:domain_decompostion_burgers}
\end{figure}

First, we evaluate the performance of the proposed hypernetwork architecture (FB-HyPINN) tasked to perform finite basis domain decomposition for physics-informed learning. In particular, we investigate two highly nonlinear case studies 1) high-frequency sinusoidal wave and 2) 1D viscous Burgers' equation. We compare the performance of FB-HyPINN with vanilla FBPINN as the baseline model. The results are presented in Table \ref{fb-hypinn:results} and Figure \ref{fig:domain_decompostion_burgers}. We used 28 subdomains and 25 subdomains for the sinusoidal and Burgers' equations respectively. We achieve comparable performance for the sinusoidal case with 23\% less trainable parameters and outperform the baseline model in the Burgers' equation case. Figure \ref{fig:domain_decompostion_burgers} demonstrates that in comparison to FBPINN, our model was able to more accurately predict the solution near the discontinuity. Details regarding the experiments and model architectures are provided in \ref{appendix : Hyperparameters} and \ref{appendix: baseline}.

\begin{table} [ht]
    \caption{Comparitive analysis on 1D Sinusoidal and 1D viscous Burgers' equations for FBPINN and FB-HyPINN. Mean relative \(L_2\) error (Rel. \(L_2\)) and mean absolute error (MAE) are reported for each method. Size refers to the total number of trainable parameters.}
    \centering
    \begin{tabular}  {c |c c c |c c c}
        \toprule
        \multirow{3}{*}{Model} 
        & \multicolumn{3}{c}{1D Burgers'} & \multicolumn{3}{c}{1D Sinusoidal} \\ \cmidrule(lr){2-4} \cmidrule(lr){5-7}
        & \multirow{2}{*}{Size} & Rel. $L_2$  & MAE & \multirow{2}{*}{Size} & Rel. $L_2$ & MAE\\ 
        & & ($\times 10^{-7}$) & ($\times 10^{-2}$) & & ($\times 10^{-2}$) & ($\times 10^{-2}$) \\
        \midrule
        FBPINN \cite{moseley2023finite}      & 8425 & 2.3 & 5.1 & 8988 & 1.07 & 5.4       \\ \cmidrule(lr){1-7} 
        \textbf{FB-HyPINN (ours)} & 5221   & 0.36 & 1.02 &  6865   & 3.45 &  5.39       \\
        \bottomrule
    \end{tabular}
    \vspace{0.1cm}
    \label{fb-hypinn:results}
\end{table}

\subsection{Learning solution operators with parameter-efficient FB-HyDON} 

We evaluate the performance of our model on three benchmark problems, namely, harmonic oscillator, 1D Burgers' equation and 1D Allen-Cahn equation, and compare it against three physics-informed baselines, DeepONet (DON) \cite{lu2021learning}, modified DeepONet (MDON) \cite{wang2021learning} and HyperDeepONet (HDON) \cite{lee2023hyperdeeponet}. For all cases, the task is to learn the solution operator mapping the initial condition \(a\) to the solution \(u\). For Burgers' and Allen-Cahn equations, we implement two variants of our model, one with a single hypernetwork (FB-HyDON-1) which takes in both input function and subdomain information as input, and one with a dual-hypernetwork module (FB-HyDON-2). Details of baseline models’ architecture and training as well as the benchmark problems' setting are provided in \ref{appendix:model details} and \ref{appendix: baseline}. 

\begin{figure}[ht] 
    \centering
    \includegraphics[width=\linewidth]{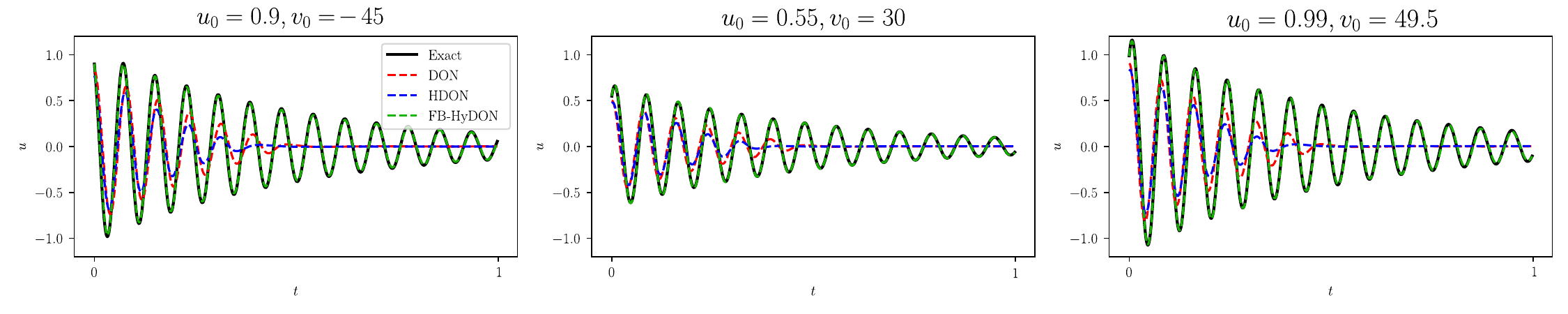} 
    \caption{Comparison of operators' prediction for solving the high-frequency harmonic oscillator with \(w_0 = 80\) via physics-informed training. Each subplot represents a unique initial condition pair \([u_0, v_0]\). Both DON and HDON failed to learn the solution operator, while FB-HyDON was able to capture the system's nonlinearities and accurately predict the solution leveraging its domain decomposition capabilities. 12 subdomains were used to train FB-HyDON.}
    \label{fig:oscillator-results}
\end{figure}

\textbf{High-frequency harmonic oscillator}. We considered a case of high-frequency harmonic oscillator (\(\omega_0 = 80\)) and trained the operator to learn the mapping between the initial condition (various pairs of initial displacement and initial velocity \([u_0, v_0]\)) and the mass displacement over time. As illustrated in Figure \ref{fig:oscillator-results}, both the DON and HDON models were unsuccessful in learning the solution operator and capturing the high-frequency characteristics of the system. In contrast, the FB-HyDON model effectively harnessed its domain decomposition capabilities to accurately predict the solution and fully encompass the system's nonlinearities. For this example, 12 subdomains were used for the training of FB-HyDON.

\begin{figure}[ht] 
    \centering
    \includegraphics[width=\linewidth]{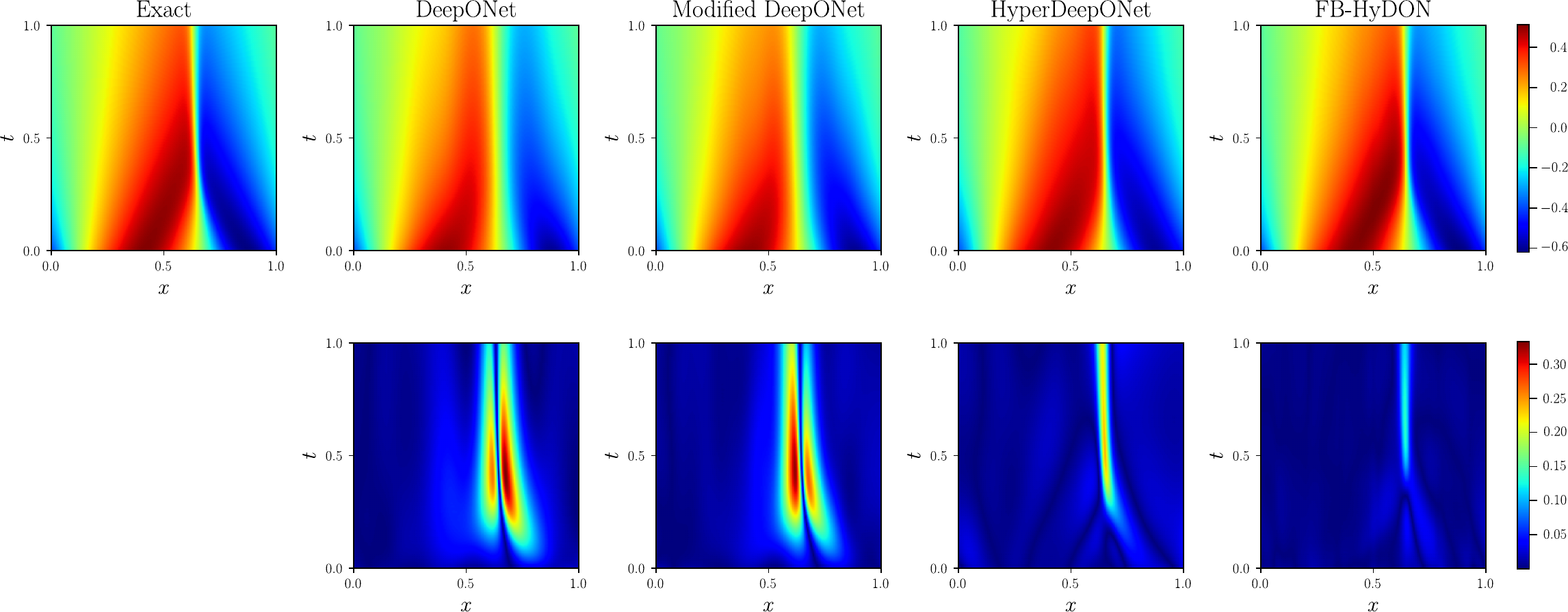} 
    \caption{Comparison of operators' prediction (top row) and absolute error (bottom row) for solving 1D Burger's equation with \(\nu = 0.005\) via physics-informed training. DeepONet and its modified version produced the largest error while having more trainable parameters. FB-HyDON (ours) outperformed other models and generated the most accurate predictions.}
    \label{fig:burgers-results}
\end{figure}

\textbf{1D Burgers' equation}. Figure \ref{fig:burgers-results} and Table \ref{PINO_table:burgers} compare the performance of the models for 1D Burgers' equation with high and low viscosity values. Both variants of our models significantly outperform other methods resulting in at least a 36\% percent reduction in Relative \(L_2\) error compared to the next best performer, HyDON. Our model also achieves remarkable efficiency in terms of network size. With only 95k parameters in the high viscosity case, it matches the compact architecture of HyDON, while considerably outperforming the larger DON (215k parameters) and MDON (203k parameters). Additionally, we observe that utilizing the dual-hypernetwork module improved the model's generalization performance while only marginally increasing the network size. For lower viscosity (higher nonlinearities and sharper transitions), unlike other methods, our models are able to maintain their high performance thanks to the built-in domain decomposition feature. By increasing the number of subdomains while keeping the number of parameters constant, FB-HyDON can successfully learn the problem's complexities.

\begin{table} 
    \caption{Comparitive analysis on 1D viscous Burgers' equations at high and low viscosity values and 1D Allen-Cahn equation. FB-HyDON-1 and FB-HyDON-2 are trained using 4 subdomains for \(v = 0.01\), 16 subdomains for \(v = 0.005\) and 8 subdomains for Allen-Cahn equation.}
    \centering
    \begin{tabular}  {c |c c |c c |c c}
        \toprule
        \multirow{3}{*}{Model} 
        & \multicolumn{2}{c}{\textbf{Burgers} (\(\nu = 0.01\))} & \multicolumn{2}{c}{\textbf{Burgers} (\(\nu = 0.005\))} & \multicolumn{2}{c}{\textbf{Allen-Cahn}} \\ \cmidrule(lr){2-3} \cmidrule(lr){4-5}\cmidrule(lr){6-7}
        & \multirow{1}{*}{Size} & Rel. $L_2$ & \multirow{1}{*}{Size} & Rel. $L_2$  & \multirow{1}{*}{Size} & Rel. $L_2$ \\
        \midrule
        DON \cite{lu2021learning}    &  215k   & 0.14  &  215k  &  0.23   &  278k & 0.93       \\
        MDON \cite{wang2021learning}    & 203k   & 0.11  & 203k   & 0.19  &  291k & 0.91          \\
        HyDON \cite{lee2023hyperdeeponet}   &  95k   & 0.089  & 128k    & 0.11  & 229k & 0.87          \\ \cmidrule(lr){1-7}
        \textbf{FB-HyDON-1 (ours)} & 95k   & 0.057 &  128k   & 0.063  &  197k & 0.24          \\ 
        \textbf{FB-HyDON-2 (ours)} & 99k   & \textbf{0.048}  &  130k   &   \textbf{0.051 }  & 202k & \textbf{0.17}   \\
        \bottomrule
    \end{tabular}
    \vspace{0.1cm}
    \label{PINO_table:burgers}
\end{table}

\begin{figure}[ht] 
    \centering
    \includegraphics[width=\linewidth]{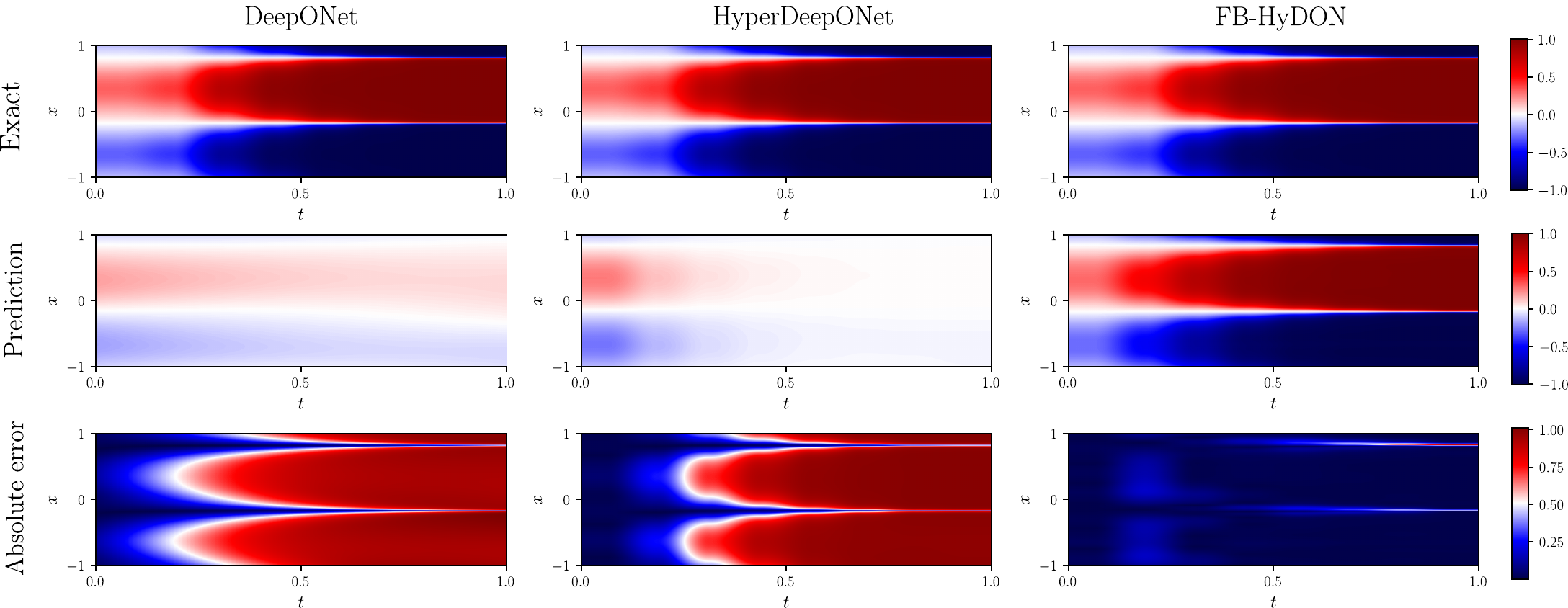} 
    \caption{Comparison of operators' prediction (middle row) and absolute error (bottom row) for solving the Allen-Cahn equation with \(\omega = 0.0001\) via physics-informed training. The ground truth solution is depicted on the top row. FB-HyDON-2 is trained using 8 subdomains.}
    \label{fig:ac-results}
\end{figure}

\textbf{1D Allen-Cahn equation}. We evaluate the effectiveness of our proposed architecture in addressing the nonlinearities and stiff terms characteristic of phase-field equations, particularly focusing on the Allen-Cahn equation. The results are provided in Figure \ref{fig:ac-results} and Table \ref{PINO_table:burgers}. Our analysis shows that domain decomposition is crucial for accurately learning the system's underlying behavior. Both the DON and HDON models exhibit poor performance, significantly deviating from the ground truth and failing to capture the overall trend. In contrast, FB-HyDON demonstrates considerably superior performance, achieving accurate predictions throughout the domain.

\section{Conclusion}
This paper introduced FB-HyDON, a novel operator architecture that addresses key limitations in existing physics-informed operators. By incorporating a built-in domain decomposition feature via finite basis functions and hypernetwork, our model achieves improved parameter efficiency and enhanced performance, particularly for highly nonlinear systems. Validation on the high-frequency harmonic oscillator, 1D Burgers' and Allen-Cahn equations demonstrated substantial improvements over state-of-the-art operators. FB-HyDON's ability to increase the number of subdomains without increasing the parameter count allows it to learn and represent complex dynamics across various scales effectively.

\bibliographystyle{plain}
\bibliography{references} 

\appendix
\clearpage

\section{Additional results}

\textbf{Effect of the number of subdomains}. Our results reveal a nuanced relationship between the number of subdomains and FB-HyDON's performance, given a fixed parameter count, as shown in Table \ref{table:size effect}. Initially, increasing the subdomain count leads to consistent performance improvements. However, this trend eventually reverses, with performance degrading beyond a certain threshold (here, 16 subdomains). This means that finer subdomain partitioning does not necessarily lead to model improvement. We hypothesize that there are two main factors contributing to this phenomenon. First, as the number of subdomains grows, the hypernetwork requires greater capacity to effectively manage the intricacies of each subdomain. With a fixed parameter budget, this increased demand may not be adequately met. Second, in configurations with numerous subdomains, the flow of information from the boundary and initial subdomains to central regions becomes increasingly challenging, which can result in suboptimal performance, as observed in \cite{dolean2024multilevel}.
\begin{table}[h]
\centering
\begin{tabular}{c|c c c c}
\hline
\# of Subdomains & 4 & 8 & 16 & 32 \\ \hline
Rel. \(L_2\)    & 0.0486     & 0.0482     & \textbf{0.0416 }     & 0.0534      \\ 
MAE        & 0.0343     & \textbf{0.0341}     & 0.0369      & 0.0392      \\ \hline
\end{tabular}
\caption{Effect of number of subdomains on the performance of FB-HyDON.}
\label{table:size effect}
\end{table}

\section{Model implementation and experiments details} \label{appendix:model details}
\textbf{Physics-informed domain decomposition experiments}.
The details of FB-HyPINN and FBPINN architectures are given in Table \ref{fbhypinn_model_config} and \ref{fbpinn_model_config} respectively. Both models are implemented in JAX \cite{jax2018github}. For FBPINN, we developed a version of the model taking FBPINN's open-source code repository \footnote{https://github.com/benmoseley/FBPINNs} as reference.

\begin{table} [ht]
    \caption{Architecture of FB-HyPINN model used for training on sinusoidal wave and 1D Burgers' equation.}
    \centering
    \begin{tabular}{c c c c c c c }
    \hline
    \textbf{Problem} & \textbf{Number of Subdomains} & \textbf{Hypernet} & \textbf{AF} & \textbf{Target Net} & \textbf{AF} & \textbf{Params} \\ 
    \hline
    Sinusoidal & 28 & \([16] \times 6\) & ReLU & \([16] \times 2\) & tanh & 6865 \\
    \hline
    {1D Burgers} & 25 & \([12] \times 6\) & ReLU & \([16] \times 2\) & tanh & 5221 \\ 
    \hline
    \end{tabular}
    \label{fbhypinn_model_config}
\end{table}

\begin{table} [ht]
    \caption{Architecture of the baseline FBPINN model used for training on sinusoidal wave and 1D Burgers' equation.}
    \centering
    \begin{tabular}{c c c c c c }
    \hline
    \textbf{Problem} & \textbf{Number of Subdomains} & \textbf{AF} & \textbf{Target Net} & \textbf{AF} & \textbf{Params} \\ 
    \hline
    Sinusoidal & 28 &  ReLU & \([16] \times 2\) & tanh & 8988 \\
    \hline
    {1D Burgers} & 25 & ReLU & \([16] \times 2\) & tanh & 8425 \\ 
    \hline
    
    \end{tabular}
    \label{fbpinn_model_config}
\end{table}

For the high-frequency sinusoidal case, we trained both models for 50k epochs. Both models were trained using Adam optimizer with a learning rate of $1 \times 10^{-3}$ and a decay rate of 0.95 per 1000 epochs. We randomly sample 200 collocation points from the input domain for training and 1000 equally spaced points for testing. The domain was decomposed into 28 overlapping subdomains with a subdomain length of 0.47. The frequency $\omega$ of the sinusoidal function was taken as 15. 

For 1D Burgers' equation, we trained both models for 50k epochs. Both models were trained using Adam optimizer with a learning rate of $1 \times 10^{-3}$ and a decay rate of 0.95 per 5000 epochs. For training randomly we sample 40,000 collocation points, 200 BC points, and 200 IC points from the input domain. Testing was done on 160,000 equally spaced points along both input dimensions (\(400 \times 400\)). The domain was decomposed into 5 subdomains along the spacial dimension and 5 across the temporal dimension giving a total of 25 overlapping subdomains. The length of the subdomains was selected as 0.35 along temporal dimension and 0.7 across the spacial dimension

\textbf{Operator learning experiments}. Details regarding the architectures of FB-HyDON-1 and FB-HyDON-2 are provided in Table \ref{our_model_config}. Both models are developed in JAX \cite{jax2018github}. We implemented the baseline models using the open-source implementation of physics-informed DeepONet and modified DeepONet \footnote{https://github.com/PredictiveIntelligenceLab/Physics-informed-DeepONets}. For HDON, we developed a physics-informed implementation based on the paper \cite{lee2023hyperdeeponet} in JAX.

\begin{table} [ht]
    \caption{Architecture of FB-HyDON used for training on harmonic oscillator and 1D Burgers' equation. 1 and 2 in the model name refer to the number of hypernetworks used in the model architecture.}
    \centering
    \begin{tabular}{c c c c c c c }
    \hline
    \textbf{Problem} & \textbf{Model} & \textbf{Hypernet} & \textbf{AF} & \textbf{Target Net} & \textbf{AF} & \textbf{Params} \\ 
    \hline
    Oscillator & FB-HyDON-1 & \([32] \times 5\) & ReLU & \([16] \times 2\) & tanh & 6.5k \\
    \hline
    {1D Burgers} & FB-HyDON-1 & \([90] \times 6\) & ReLU & \([32] \times 4\) & tanh & 95k \\
      \(\nu = 0.01\)  & FB-HyDON-2 & \([64] \times 5\) \(+\)  \([64] \times 3\) & ReLU & \([32] \times 4\) & tanh & 99k \\
    \hline
        {1D Burgers} & FB-HyDON-1 & \([100] \times 6\) & ReLU & \([32] \times 5\) & tanh & 128k \\
      \(\nu = 0.005\)  & FB-HyDON-2 & \([64] \times 5\) \(+\)  \([64] \times 5\) & ReLU & \([32] \times 5\) & tanh & 130k \\
      \hline
       \multirow{2}{*}{1D Allen-Cahn}& FB-HyDON-1 & \([100] \times 6\) $ $ & ReLU & \([45] \times 5\) & 
       tanh & 197k \\
       & FB-HyDON-2 & \([90] \times 5\) \(+\)  \([90] \times 5\) & ReLU & \([32] \times 5\) & tanh & 202k \\

      \hline
    \label{our_model_config}
    \end{tabular}
\end{table}

\begin{table} 
    \caption{Architecture of baseline models used for training on harmonic oscillator and 1D Burgers' equation.}
    \centering
    \begin{tabular}{c c c c c c c }
    \hline
    \textbf{Problem} & \textbf{Model} & \textbf{Branch/hypernet} & \textbf{AF} & \textbf{Trunk/target Net} & \textbf{AF} & \textbf{Params} \\ 
    \hline
    \multirow{3}{*}{Oscillator} & DON & \([32] \times 4\) & tanh & \([32] \times 4\) & tanh & 7.7k \\
                                   & MDON & \([32] \times 4\) & tanh & \([32] \times 4\) & tanh & 10k  \\
                                   & HDON & \([32] \times 5\) & ReLU & \([16] \times 2\) & tanh & 6.1k \\
    \hline
    \multirow{2}{*}{1D Burgers} & DON & \([128] \times 7\) & tanh & \([128] \times 7\) & tanh & 203k \\
                                & MDON & \([128] \times 7\) & tanh & \([128] \times 7\) & tanh & 215k \\
                                \(\nu = 0.01\) & HDON & \([90] \times 6\) & ReLU & \([32] \times 4\) & tanh &
                                95k \\
    \hline
    \multirow{2}{*}{1D Burgers} & DON & \([128] \times 7\) & tanh & \([128] \times 7\) & tanh & 203k \\
                            & MDON & \([128] \times 7\) & tanh & \([128] \times 7\) & tanh & 215k \\
                            \(\nu = 0.005\) & HDON & \([100] \times 6\) & ReLU & \([32] \times 5\) & tanh &
                            128k \\
    \hline
        \multirow{3}{*}{1D Allen-Cahn} & DON & \([150] \times 7\) & tanh & \([150] \times 7\) & tanh & 278k \\
                            & MDON & \([150] \times 7\) & tanh & \([150] \times 7\) & tanh & 291k \\
                             & HDON & \([100] \times 6\) & ReLU & \([50] \times 5\) & tanh &
                            229k \\
    \hline
    \label{our_model_config}
    \end{tabular}
\end{table}

For the harmonic oscillator, the models were trained for 100k iterations on the full batch of data. We sampled 200 initial condition pairs \([u_0, v_0]\) for training and three for testing from \([0, 1]\) for initial displacement \(u_0\) and \([-50, +50]\) for initial velocity \(v_0\). Adam optimizer with an initial learning rate of \(1 \times 10^{-3}\) and a decay rate of 0.9 per 10k epochs was used for all cases. 500 collocation points and 1 initial condition point were used to minimize the physics-informed losses.

For both 1D Burgers' and Allen-Cahn equations, we trained the models for 25k iterations with abatch size of 1000. 1000 training and 100 test input functions (initial conditions) are randomly generated using a mean-zero Gaussian random field \(\textup{GRF} \sim \mathcal{N}\left(0,25^2\left(-\Delta+5^2 I\right)^{-4}\right)\) and used for training and evaluation of the operators. 40 equally-spaced sensor points were used for encoding the input function information. All models were trained using Adam optimizer with an initial learning rate of \(1 \times 10^{-3}\) and an exponential decay rate of 0.9 per 1000 epochs. For each training input function, 2500 collocation points, 100 IC points and 100 BC points were uniformly sampled and used for training. For models with hypernetwork, we used the chunking strategy and set the number of chunks to 6. 

\label{appendix : Hyperparameters}
\section{Benchmark problems setting} \label{appendix: baseline}
\textbf{High-frequency sinusoidal}.
Following \cite{moseley2023finite}, this case represents a 1D scenario where neural networks often encounter spectral bias, necessitating a domain decomposition approach. Considering the following problem:
\[
\frac{du}{dx} = \cos(\omega x),
\]
\[
u(0) = 0,
\]

where \( x, u, \omega \in \mathbb{R} \) is the frequency of the wave. The exact solution to this problem is given by:

\[
u(x) = \frac{1}{\omega} \sin(\omega x).
\]

\textbf{1D Burgers' equation for FB-HyPINNs}.
We tested the proposed FB-HyPINN and FBPINNs on 1D viscous time-dependent Burgers equation, given by:
\[
\frac{\partial u}{\partial t} + u \frac{\partial u}{\partial x} = \nu \frac{\partial^2 u}{\partial x^2},
\]

with initial and boundary conditions,
\[
u(x, 0) = -\sin(\pi x),
\]
\[
u(-1, t) = 0,
\]
\[
u(1, t) = 0,
\]

where \( x, t, u, \nu \in \mathbb{R} \), and the problem domain is \( x \in [-1, 1] \) and \( t \in [0, 1] \). Notably, for small values of the viscosity parameter \( \nu \), specifically \( \nu = \frac{0.01}{\pi} \), the solution develops a discontinuity at \( x = 0 \) as time progresses. The exact solution can be obtained from open source code repository \footnote{https://github.com/benmoseley/FBPINNs/tree/main/fbpinns/traditional\_solutions/analytical}, that uses the Hopf-Cole transform as explained in  \cite{BASDEVANT198623}.

\textbf{Harmonic Oscillator for operator learning}. We solved the 1D damped harmonic oscillator, where we aim to model the displacement \( u(t) \) of the oscillator over time. The problem is defined by the following ordinary differential equation (ODE):

\begin{equation}
m \frac{d^2 u}{d t^2} + \mu \frac{d u}{d t} + ku = 0,
\end{equation}

where \( m \) is the mass of the oscillator, \( \mu \) is the damping coefficient, and \( k \) is the spring constant. Considering \(\quad \delta = \frac{\mu}{2m}\) and \(\quad \omega_0 = \sqrt{\frac{k}{m}}\), we investigated the under-damped state where \(\delta < \omega_0\) \cite{moseley2023finite}. With the initial conditions \(u(0) = u_0\) and \(\frac{du}{dt}(0) = v_0\) the exact solution will be:

\begin{equation}
    u(t) = e^{-\delta t} \left( u_0 \cos(\omega t) + \frac{v_0 + \delta u_0}{\omega} \sin(\omega t) \right).
\end{equation}

\textbf{1D Burgers' equation for operator learning}. To demonstrate the capability of our proposed operator in addressing nonlinearities in governing PDEs, we utilized the 1D Burgers' equation benchmark, following the setup described in \cite{li2020fourier}. The equation is given by:

\begin{equation}
\frac{\partial u}{\partial t} + u \frac{\partial u}{\partial x} - \nu \frac{\partial^2 u}{\partial x^2} = 0, \quad (x,t) \in (0,1) \times (0,1],
\end{equation}
\begin{equation}
u(x,0) = a, \quad x \in (0,1),
\end{equation}
with periodic boundary conditions:
\begin{equation}
u(0,t) = u(1,t),
\end{equation}
\begin{equation}
\frac{\partial u}{\partial x}(0,t) = \frac{\partial u}{\partial x}(1,t).
\end{equation}

In our experiments, we set the viscosity to $\nu = 0.01$ and $\nu = 0.005$ and generated the initial condition $a$ from a GRF. To generate the solution test data, we followed the procedure outlined in \cite{wang2021learning}. We solved the 1D Burgers' equation using conventional spectral methods, assuming periodic boundary conditions. The initial condition \(s(x, 0) = u(x)\) was integrated up to a final time \(t = 1\), with temporal snapshots of the solution recorded at regular intervals.

\textbf{1D Allen-Cahn for operator learning}. For 1D Allen-Cahn equation with periodic boundary conditions, we followed the setup described in \cite{raissi2019physics}:

\begin{equation}
\frac{\partial u}{\partial t} - 0.0001\frac{\partial^2 u}{\partial x^2} + 5u^3 - 5u = 0, \quad x \in [-1,1], \quad t \in [0,1],
\end{equation}
\begin{equation}
u(x, 0) = a,
\end{equation}
\begin{equation}
u(-1, t) = u(1, t),
\end{equation}
\begin{equation}
\frac{\partial u}{\partial x}(-1, t) = \frac{\partial u}{\partial x}(1, t).
\end{equation}

\end{document}